\RequirePackage[table]{xcolor}
\documentclass[journal,twoside,web]{ieeecolor}

\usepackage{generic}
\usepackage{cite}
\usepackage{amsmath,amssymb,amsfonts}
\usepackage{algorithm,algorithmic}
\usepackage{graphicx}
\usepackage[hidelinks]{hyperref}
\usepackage{textcomp}
\usepackage{booktabs}
\usepackage{tabularx}
\usepackage[caption=false,font=footnotesize]{subfig}

\newtheorem{claim}{Claim}

\definecolor{exactcol}{RGB}{204,229,255}
\definecolor{uomcol}{RGB}{255,218,185}

\makeatletter
\def\ps@titlepagestyle{\def\@oddfoot{}\def\@evenfoot{}%
  \def\@oddhead{\begin{tabular}{@{}p{43pc}@{}}\\[-20pt]\vbox{\hsize43pc\scriptsize\textsf\leftmark \hfill \textsf\thepage\hbox{}\par\vspace*{-3pt}
    \vbox{\color{subsectioncolor}\hrule height1pt width43pc depth0pt}}\end{tabular}}%
  \def\@evenhead{\@oddhead}}
\makeatother

\begin{document}

\title{Peer-Grounded Counterfactual Path Planning for Chronic Health Management}

\author{Saman Khamesian\textsuperscript{1,2}* and Hassan Ghasemzadeh\textsuperscript{1}
\thanks{$^{1}$College of Health Solutions, Arizona State University, Phoenix, USA}
\thanks{$^{2}$School of Computing and Augmented Intelligence, Arizona State University, Tempe, USA}
\thanks{{*\textbf{Corresponding author}: \textcolor{blue}{skhamesi@asu.edu}}}}

\maketitle

\begin{abstract}
Effective behavioral intervention in chronic disease management requires a sequence of incremental steps, each grounded in what similar individuals have achieved. Multimodal records of daily behavior make such evidence available outside the clinic. Counterfactual explanation, an explainable AI approach, can turn these records into guidance, yet endpoint-only methods return a target state without a route to it. We propose \textit{POROS} (\underline{P}eer-Grounded \underline{O}ptimal \underline{R}outes \underline{O}ver \underline{S}tates), a domain-agnostic framework rooted in Bandura's self-efficacy theory and Festinger's social comparison theory. POROS constructs a Behavioral Progression Graph---a directed acyclic graph over observed patient states in which every edge requires both peer-grounded behavioral proximity and strict improvement in the observed health outcome. Every edge is bounded by a feasibility threshold derived from the cohort's day-to-day behavioral changes. Minimum-cost paths through this graph decompose behavioral gaps into incremental, peer-grounded steps. We evaluate POROS on two independent longitudinal cohorts of patients with type 1 diabetes, with a per-patient predictor estimating individual time in range (TIR, blood glucose within 70--180~mg/dL) along each path. Compared with an endpoint-only baseline, POROS returns multi-hop paths to the same target for 14.4\% and 19.8\% of days below the 70\% TIR threshold, reducing the per-step change in carbohydrate intake, insulin dose, and bolus timing by 38--42\% on the larger cohort. POROS also supports improvement within the same outcome class, connecting 78.1\% and 83.7\% of outcome-ordered pairs of well-controlled days. Together, these findings demonstrate how POROS translates multimodal health records into incremental, peer-grounded guidance for chronic disease management through observational path planning.
\end{abstract}


\section{Introduction}
\label{sec:introduction}

\IEEEPARstart{M}{any} chronic health conditions are managed not through a single clinical event but through the cumulative effect of daily behavioral choices. A patient managing their weight adjusts what and how much they eat, how frequently they are active, and when they sleep. A person with Type~1 diabetes (T1D) coordinates carbohydrate intake, insulin dosing, meal timing, and correction behavior across every waking hour~\cite{ada2024,snaith2026,cummings2022}. 

Daily behavioral choices are now recorded outside the clinic by complementary multimodal data streams, ranging from wearable devices such as smartwatches that track physical activity and sleep to condition-specific records~\cite{fraser2025,inaishi2023}. In T1D, for instance, meal logs capture carbohydrate intake and meal times~\cite{aiello2022eating}, and insulin delivery records capture each dose~\cite{berget2020,sherr2023,ada2026tech}. When these modalities are aligned on a shared timeline, they reveal behavior that no single stream contains, such as how far an insulin bolus precedes or follows the meal it covers~\cite{peters2017}, or how meal timing relates to the physical activity that follows it~\cite{riddell2017,monttblanchard2023,dhaliwal2026}. Together, these multimodal records describe what a patient did each day and how much that behavior changed from one day to the next~\cite{aiello2022eating}. Yet describing behavior is not the same as guiding it, and translating these records into guidance that patients can follow between clinic visits remains central to scalable, long-term chronic care~\cite{zhu2021,khamesian2026guide}.

The central challenge of behavioral intervention design is therefore not simply identifying that behavior must change, but specifying \emph{how} to get from here to there: a sequence of steps, each small enough to be genuinely attempted, that together carry a patient from their current state toward a better one~\cite{hampson2000}. In T1D, this challenge is especially acute. Clinical guidelines define a Time in Range (TIR)---the fraction of the day blood glucose remains within 70--180~mg/dL---of at least 70\% as the threshold for adequate glycemic control~\cite{battelino2019,ada2024}, yet patients with TIR well below this target face a behavioral gap so wide that a single corrective prescription is rarely sufficient or actionable~\cite{foster2019,ebekozien2023}.

Two foundational theories from behavioral science make clear why a recommendation's structure matters as much as its content. Bandura's self-efficacy theory~\cite{bandura1997} holds that willingness to attempt behavioral change is calibrated to the magnitude of the next required step, not the distance to the final goal---so a recommendation demanding simultaneous large adjustments across multiple dimensions may be mathematically optimal yet psychologically inert. Festinger's social comparison theory~\cite{festinger1954} adds a complementary mechanism: people gauge what is achievable by observing similar others, making a concrete peer-demonstrated profile more motivating than an abstract target. Together they converge on a structural requirement for effective behavioral guidance---recommendations must be incremental, and each step grounded in behavior that real, similar individuals have demonstrated.

\begin{figure}[t]
\centering
\includegraphics[width=\linewidth]{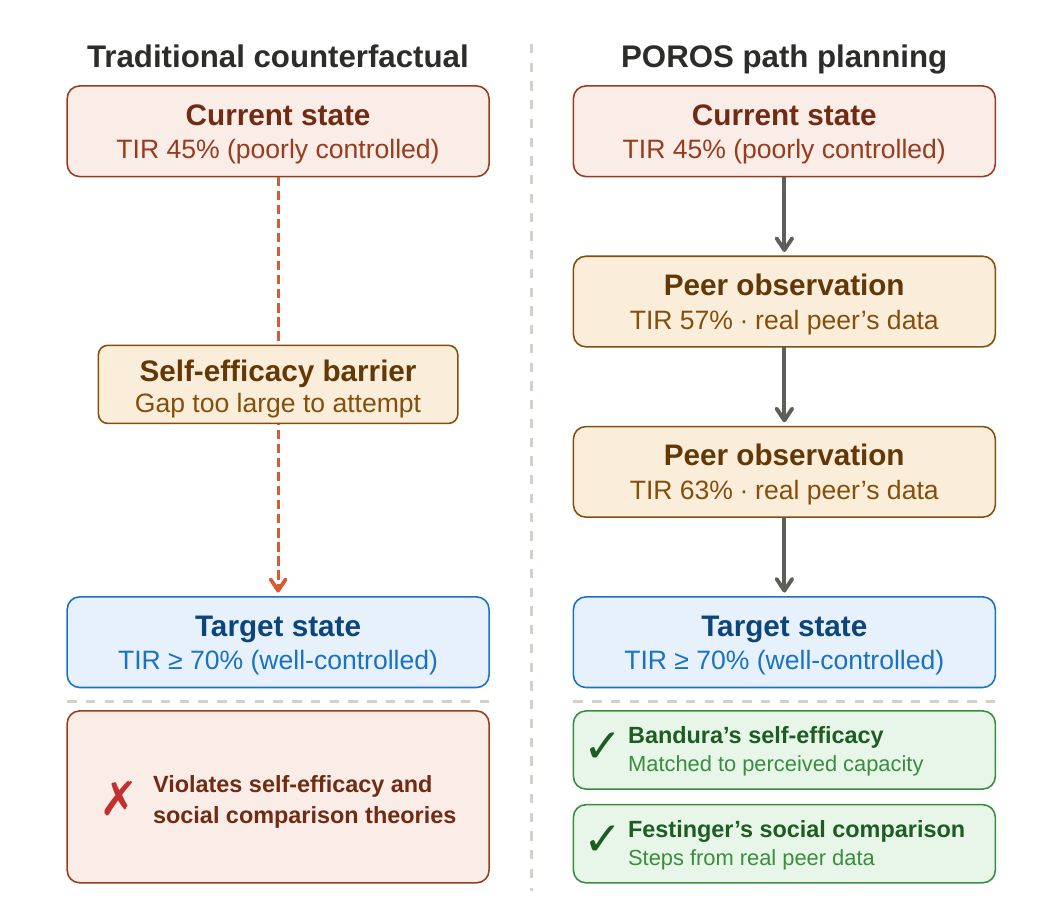}
\caption{Traditional counterfactual explanation versus POROS path planning, instantiated on type 1 diabetes management. Traditional methods return a single target state whose behavioral gap violates self-efficacy. In contrast, POROS decomposes the same gap into incremental, peer-grounded steps that embed Bandura's self-efficacy and Festinger's social comparison theories.}
\label{fig:poros_diagram}
\end{figure}

Counterfactual explanation (CE), a central approach in explainable artificial intelligence (XAI)~\cite{miller2019explanation,guidotti2024counterfactual,verma2024counterfactual, arefeen2024glyman}, offers a compelling computational framework for generating such guidance. Given an individual's current behavioral state and a model of health outcomes, a CE answers the question: what minimal change in behavior would produce a better outcome? Whereas feature-attribution methods report which inputs drove a prediction~\cite{ribeiro2016should,lundberg2017unified}, a CE tells the individual what to change, which makes it the form of explanation a patient can act on. Landmark methods such as those of Wachter et al.~\cite{wachter2017} and DiCE~\cite{mothilal2020} have established CE as a principled approach to actionable algorithmic recourse, and their application to health contexts has produced genuinely useful recommendations~\cite{lenatti2022novel,arefeen2025glytwin}---in T1D, for instance, a CE can translate directly into a behavioral prescription specifying adjustments in insulin dosing, carbohydrate intake, and meal timing that would have resulted in a well-controlled day.

The fundamental limitation, however, is that nearly all CE methods produce a \emph{destination}, not a roadmap---a single target state, presented without any account of how to traverse the gap between where the individual is and where the CE says they should be. For patients far from their health target, telling them where to end up without a sequence of feasible intermediate steps is precisely the recommendation structure that self-efficacy theory predicts will go unattempted. Path-aware extensions use data density to select a more reachable endpoint~\cite{poyiadzi2020,small2023}, yet still return that endpoint rather than the path, optimizing for minimal behavioral change in feature space with no guarantee that the health outcome improves monotonically along the way. The result is a trajectory that may be mathematically smooth but remains disconnected from the lived reality of behavioral change.

We propose \textbf{POROS} \textit{(\underline{P}eer-Grounded \underline{O}ptimal \underline{R}outes \underline{O}ver \underline{S}tates)}, addressing these limitations in a unified way. POROS constructs a Behavioral Progression Graph---a directed acyclic graph (DAG) over patient-day states in which an edge exists only if the transition is peer-grounded and the health outcome strictly improves. Each node combines the multimodal behavioral records of one observation period into a single profile, paired with its measured health outcome. Rather than synthesizing a hypothetical target state, POROS uses these real observations to find paths through behavioral profiles that individuals in the cohort have actually exhibited. POROS is therefore an observational, peer-grounded trajectory discovery framework. Figure~\ref{fig:poros_diagram} illustrates this contrast for the T1D instantiation. We summarize our contributions as follows.

\begin{itemize}
    \item \textbf{Peer-grounded edges.} A transition is admitted only when the behavioral distance between two states stays within a threshold derived from the cohort's own within-subject behavioral variation, so no edge is larger than the biggest single-period change that every individual in the cohort has made. Because edges form with no restriction on individual identity, every step corresponds to a behavioral profile drawn from a real peer, structurally embedding social comparison into each recommendation POROS produces.

    \item \textbf{Incremental paths.} Minimum-cost paths through the graph decompose what would otherwise be a single large behavioral prescription into a sequence of incremental steps, each small enough to support the self-efficacy required for genuine behavioral attempt.

    \item \textbf{A clinically grounded distance metric.} We introduce the Clinical Behavioral Transition Distance (CBTD), which rescales each behavioral feature by its minimum clinically important difference (MCID)~\cite{jaeschke1989} so that one unit of distance corresponds to one clinically meaningful step. This places features from different modalities, measured in grams, units and minutes, on a single clinically meaningful scale.

    \item \textbf{Monotone improvement by construction.} Every edge requires a strict gain in the observed health outcome, so the observed outcome rises at every step of every path regardless of how that path is found, rather than being evaluated at the endpoint alone.

    \item \textbf{Query-agnostic construction.} Once constructed, the same graph supports path reconstruction, reachability analysis, and peer identification without retraining. No classifier appears anywhere in construction, so a query is not restricted to crossing between classes and any transition toward a better health outcome is supported.
\end{itemize}

\section{Related Work}
\label{sec:related_work}

\subsection{Counterfactual Explanations (CE)}
\label{sec:counterfactual_explanations}

Counterfactual explanation was formalized by Wachter et al. \cite{wachter2017} as the minimal feature-space perturbation that flips a classifier's output to a desired class. The output is a single alternative instance --- an endpoint --- with no trajectory or sequence connecting it to the original. DiCE \cite{mothilal2020} extended this formulation by generating a diverse set of such endpoints through determinantal-point-process optimization, improving coverage of the counterfactual space while retaining the same static, point-based output structure. Both methods are model-agnostic and have proven broadly influential, but they share the structural property that defines their limitation in behavioral health contexts: the recommendation is where the individual should end up, not how to get there.

\subsection{Graph-Based, Path-Aware, and Sequential CE}
\label{sec:graph_based_ce}

Several methods go beyond a single endpoint and describe how to reach a counterfactual. Graph-based methods such as FACE~\cite{poyiadzi2020}, LocalFACE~\cite{small2023}, and the fairness-auditing approach of Fragkathoulas et al.~\cite{fragkathoulas2026facegroup} construct a graph over observed instances and route through high-density regions of the data via shortest-path search. Sequential recourse methods such as OrdCE~\cite{kanamori2021ordered} and FastAR~\cite{verma2022amortized} instead order the feature changes of an action or learn a policy that moves an instance through intermediate states, and these states are synthesized rather than observed. In all of these methods, the destination is a counterfactual defined by a trained classifier's decision boundary. Yet flipping a classifier's decision does not guarantee real improvement. Improvement-focused causal recourse~\cite{konig2023improvement} addressed this gap by guiding toward improvement in the real-world state, which it achieves only given correct causal knowledge.

POROS differs from these methods in a more fundamental way. An edge in its Behavioral Progression Graph is not defined by any decision boundary. It requires only that a transition be peer-grounded and that a measured health outcome strictly improves. Every intermediate state is therefore a real observation, and no classifier or density model is needed to connect two states. POROS pursues the same goal as improvement-focused recourse without a causal model, discovering peer-grounded trajectories from observed outcomes rather than estimating the causal effect of an intervention on the individual. Because outcome improvement is continuous rather than a binary label, edges form even within the same outcome class. In T1D, for instance, an edge can represent a rise in TIR from 45\% to 60\%, a clinically meaningful gain that never crosses the 70\% threshold and is invisible to methods built around crossing a decision boundary. The graph therefore supports a family of queries over any pair of outcome-ordered states rather than a single counterfactual target.

\subsection{CE for Behavioral Health Management}
Recent work has brought CE into glycemic management for T1D. GlyMan \cite{arefeen2024glyman} generates counterfactual behavioral recommendations for patients on automated insulin delivery systems, incorporating patient preferences into the generation process. GlyTwin \cite{arefeen2025glytwin} extends this into a digital twin framework, using counterfactual explanation to simulate optimal behavioral treatments and reduce hyperglycemic events. Both methods follow the endpoint framing of their predecessors: the recommendation specifies where the patient should be, with no prescribed sequence for reaching that state. POROS departs from this class in that the path is the recommendation, every node along it is a real observed patient-day, and the observed TIR strictly increases at every step by construction rather than being evaluated at the endpoint alone.

\begin{figure*}[!tp]
\centering
\includegraphics[width=\linewidth]{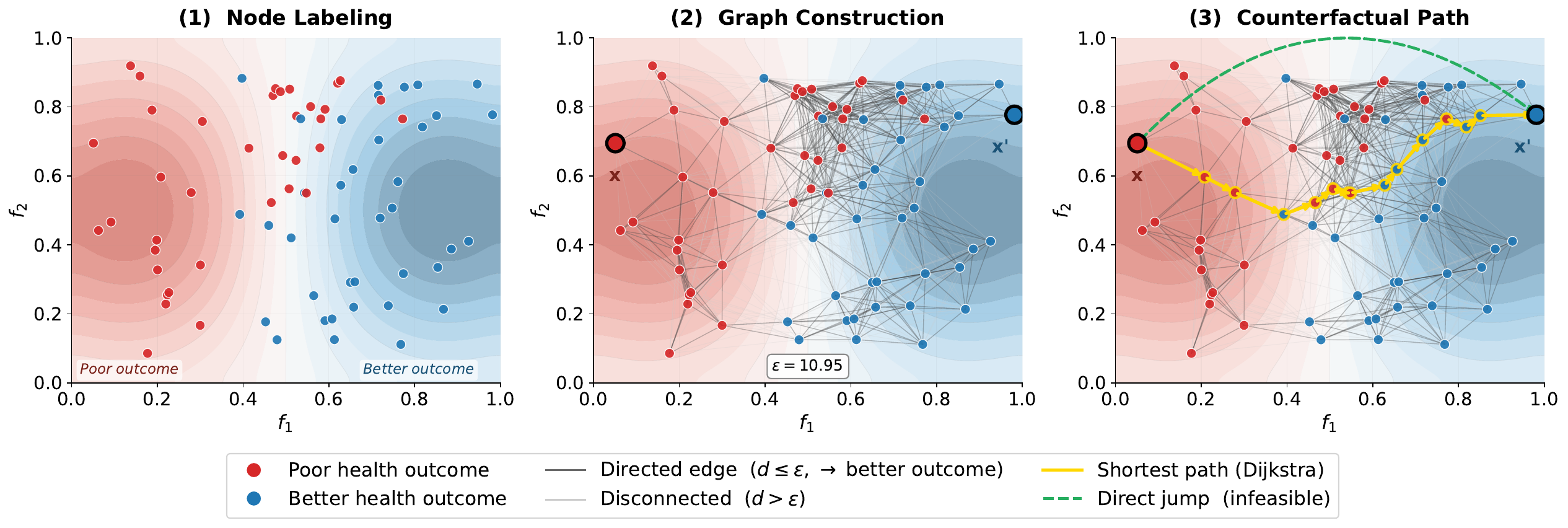}
\caption{POROS methodology and the Behavioral Progression Graph it constructs over a study cohort. Each node is one behavioral observation, colored by health outcome from red (poor) to blue (better), positioned by two behavioral features $f_1$ and $f_2$. \textbf{(1)~Node Labeling:} observations are labeled relative to a clinical threshold $\tau$. \textbf{(2)~Graph Construction:} a directed edge $x_i \to x_j$ is admitted under Eq.~\eqref{eq:adjacency}, requiring $d(\mathbf{v}_i, \mathbf{v}_j) \leq \varepsilon$ and strict outcome improvement, so every connection points toward better health. \textbf{(3)~Counterfactual Path:} Dijkstra's algorithm recovers the minimum-cost path from source $x$ to target $x'$, traversing peer observations as incremental steps. The dashed green arc is the infeasible direct transition ($d > \varepsilon$), which the graph decomposes into a multi-hop peer-grounded path.}
\label{fig:poros_methodology}
\end{figure*}

\section{Problem Statement}
\label{sec:problem_statement}

Let $V = \{x_1, \dots, x_M\}$ denote the set of observations in a cohort, where each observation captures the behavior of one individual over one observation period. Each observation $x_i \in V$ is described by a behavioral feature vector $\mathbf{v}_i \in \mathbb{R}^K$ and an observed outcome $H(x_i)$, where a higher value of $H$ indicates a more desirable outcome. Given a source observation $x \in V$ and a target observation $x' \in V$ with $H(x') > H(x)$, the goal is to recommend how the individual at $x$ can reach the outcome of $x'$. Endpoint-only counterfactual methods answer this question with a single jump from $x$ to $x'$. In practice, however, a large behavioral change in one step is rarely achievable for an individual. We therefore seek a sequence of gradual behavioral changes instead of a single endpoint.

Formally, a counterfactual path from $x$ to $x'$ is a sequence $\pi = (x, \dots, x')$ of $L$ steps, where $L$ is not fixed in advance. We define the feasible set $\Pi(x, x')$ as the set of all such sequences that satisfy three constraints. First, $x_i \in V$ for every $x_i \in \pi$, so every state on the path is an observed instance. Second, for every two consecutive states $x_i$ and $x_j$ on $\pi$, $d(\mathbf{v}_i, \mathbf{v}_j) \leq \varepsilon$, where $d(\mathbf{v}_i, \mathbf{v}_j)$ denotes the distance between the behavioral feature vectors $\mathbf{v}_i$ and $\mathbf{v}_j$ of two observations $x_i$ and $x_j$, and $\varepsilon$ bounds the size of a single feasible behavioral change. Third, for every two consecutive states $x_i$ and $x_j$ on $\pi$, $H(x_i) < H(x_j)$, so the outcome improves at every step. Given a criterion $J$ that assigns a score to each path, the path-finding function $F$ selects
\begin{equation}
\pi^{*} = F(x, x') = \operatorname*{arg\,min}_{\pi \in \Pi(x, x')} J(\pi).
\label{eq:problem}
\end{equation}

The constraints define which paths are admissible, and the criterion $J$ defines which admissible path is preferred. For instance, $J$ may measure the total transition cost along the path or the number of steps. Endpoint-only counterfactual methods, in contrast, restrict the search to $L = 1$.

\section{Method}
\label{sec:method}

To find such paths, POROS\footnote{\textbf{Source Code:} \url{https://github.com/SamanKhamesian/POROS}} constructs the Behavioral Progression Graph $G = (V, E, W)$ over the observed behavioral records of a cohort, where $V$ is the set of behavioral observations, $E$ the directed edges encoding peer-grounded, outcome-improving transitions, and $W$ the cost assigned to each admitted edge. Each node in $V$ represents a single behavioral observation (e.g., a patient-day in a longitudinal clinical study), carrying a behavioral feature vector and a measured health outcome. A directed edge from $x_i$ to $x_j$ is admitted only when two conditions hold jointly: the behavioral distance between $x_i$ and $x_j$ does not exceed a cohort-derived threshold $\varepsilon$, and the health outcome at $x_j$ strictly exceeds that at $x_i$, with no restriction on individual identity. Because every edge enforces strictly improving outcomes, $G$ is a DAG by construction. Once constructed, $G$ supports a family of queries without retraining, with minimum-cost paths recovered via Dijkstra's algorithm (Figure~\ref{fig:poros_methodology}). The remainder of this section defines each component of $G$ formally, then describes the path-finding function $F$.

\subsection{Graph Construction}
\label{sec:graph_construction}

\subsubsection{Nodes}

Let the cohort consist of $P$ individuals, with individual $p$ observed over $D_p$ periods, yielding $M = \sum_{p=1}^{P} D_p$ observations in total. The node set is $V = \{x_1, x_2, \ldots, x_M\}$, where each node $x_i$ corresponds to one observation. Every node carries two attributes.

\textbf{Behavioral feature vector.} The vector $\mathbf{v}_i \in \mathbb{R}^K$ encodes the behavioral profile of observation $i$ across $K$ behavioral features (e.g., total dietary carbohydrate intake, meal timing, physical activity, medication dosing). Physiological measurements and health outcome values are excluded from $\mathbf{v}_i$ by design; only quantities the individual can directly modify are included.

\textbf{Health outcome.} The function $H : V \rightarrow \mathbb{R}$ maps each node to its measured health outcome (e.g., TIR for diabetes management, BMI for weight management). Given a domain-specified clinical threshold $\tau$, nodes with $H(x_i) \geq \tau$ represent well-controlled states and those with $H(x_i) < \tau$ represent states requiring improvement. The Behavioral Progression Graph supports directed path queries between any two nodes with strictly ordered outcomes, not only across this threshold.

\subsubsection{Adjacency}

A directed edge exists from $x_i$ to $x_j$ if and only if:

\begin{equation}
    x_i \to x_j \quad \text{iff} \quad d(\mathbf{v}_i,\, \mathbf{v}_j) \leq \varepsilon \quad \text{and} \quad H(x_j) > H(x_i)
    \label{eq:adjacency}
\end{equation}

The first condition enforces behavioral proximity within threshold $\varepsilon$. The second enforces strict improvement in health outcome. No restriction on individual identity appears in~\eqref{eq:adjacency}; edges form freely between observations from any individuals in the cohort, provided both conditions are satisfied. Because every edge requires $H(x_j) > H(x_i)$, every directed path in $G$ is monotonically improving in observed health outcome --- the observed health outcome strictly increases at every step, by construction rather than by algorithmic choice. This structural guarantee holds for any path regardless of how it is found, making $G$ a DAG on which standard shortest-path search applies without modification (Proof 1, Appendix~\ref{app:proofs}).

\subsubsection{{$\varepsilon$} --- Transition Threshold}

The threshold $\varepsilon$ controls which behavioral transitions are admitted as edges. Rather than treating $\varepsilon$ as a free hyperparameter, we derive it from the cohort's observed within-individual behavioral dynamics so that the size of every admitted step is bounded by what individuals in the cohort have actually done, not by a manually tuned value.

For individual $p$ with feature vectors $\{\mathbf{v}_p^{(t)}\}_{t=1}^{D_p}$ in chronological order, define the per-individual maximum observed single-period behavioral transition:

\begin{equation}
    \varepsilon_p = \max_{t=1,\ldots,D_p-1} \; d\!\left(\mathbf{v}_p^{(t)},\, \mathbf{v}_p^{(t+1)}\right)
\end{equation}
The quantity $\varepsilon_p$ is the largest behavioral change individual $p$ has demonstrably made in a single observation period. The global threshold is then:
\begin{equation}
    \varepsilon = \min_p \; \varepsilon_p
    \label{eq:epsilon}
\end{equation}
By construction, $\varepsilon \leq \varepsilon_p$ for every individual $p$. It follows that for any admitted edge $(x_i, x_j) \in E$:
\begin{equation}
    d(\mathbf{v}_i, \mathbf{v}_j) \leq \varepsilon \leq \varepsilon_p \quad \text{for all } p \in \{1, \ldots, P\}
\end{equation}

Every admitted edge therefore represents a behavioral transition whose magnitude is within the demonstrated single-period capacity of every individual in the cohort --- not merely some (Proof 2, Appendix~\ref{app:proofs}). Because $\varepsilon_p$ is computed with the same distance $d$ that defines adjacency, the CBTD of Eq.~\eqref{eq:cbtd}, this bound applies to the combined change across all $K$ features, measured on the same scale on which every edge is admitted. The threshold $\varepsilon$ is therefore a cohort-derived feasibility bound on step magnitude, in contrast to path-aware methods in which the connectivity threshold is a manually chosen parameter~\cite{poyiadzi2020}.

\subsubsection{Clinical Behavioral Transition Distance}

The behavioral feature vector $\mathbf{v}$ aggregates measurements in heterogeneous units (e.g., carbohydrate intake in grams, time between meals in minutes, medication dose in units). A raw difference of two units carries a different meaning depending on the feature, and not every unit change is clinically actionable. For two nodes $x_i, x_j \in V$ with feature vectors $\mathbf{v}_i, \mathbf{v}_j \in \mathbb{R}^K$, where $v_{i,k}$ denotes the value of feature $k$ at observation $i$, we define the \textit{Clinical Behavioral Transition Distance (CBTD)} as:

\begin{equation}
    d(\mathbf{v}_i, \mathbf{v}_j) = \sqrt{\sum_{k=1}^{K} w_k \left[\frac{v_{i,k} - v_{j,k}}{\delta_k}\right]^2}
    \label{eq:cbtd}
\end{equation}

where $\delta_k > 0$ is the MCID for feature $k$, set as a domain-specific hyperparameter, and $w_k > 0$ is its relative importance weight. Dividing $v_{i,k} - v_{j,k}$ by $\delta_k$ rescales dimension $k$ so that one unit in the transformed space corresponds to exactly one clinically meaningful step, making all $K$ dimensions commensurable without recourse to population-level statistics. The CBTD is used identically in the adjacency condition~\eqref{eq:adjacency} and the $\varepsilon$ derivation~\eqref{eq:epsilon}.

\subsubsection{Edge Weights}

The weight of each admitted edge $(x_i, x_j) \in E$ is the squared CBTD:
\begin{equation}
    W(x_i, x_j) = d(\mathbf{v}_i, \mathbf{v}_j)^2
    \label{eq:weight}
\end{equation}
While $d$ governs the adjacency condition in~\eqref{eq:adjacency}, it is $d^2$ that is assigned as the edge weight for minimum-cost path computation. This choice is structurally motivated by the triangle inequality. Because CBTD is a metric, for any triple $x_i, x_k, x_j \in V$:

\begin{equation}
    d(\mathbf{v}_i, \mathbf{v}_j) \leq d(\mathbf{v}_i, \mathbf{v}_k) + d(\mathbf{v}_k, \mathbf{v}_j)
\end{equation}

Under a linear edge weight $d$, path costs inherit this property: any direct edge $(x_i, x_j) \in E$ is at least as cheap as any multi-hop alternative between the same endpoints. Any minimum-cost path criterion therefore always selects direct transitions wherever they exist, collapsing paths to the largest admissible single step and producing exactly the behavioral jumps that self-efficacy theory identifies as inactionable.

The quadratic weight $d^2$ breaks this property in cost space. Letting $a = d(\mathbf{v}_i, \mathbf{v}_k)$ and $b = d(\mathbf{v}_k, \mathbf{v}_j)$, no such bound holds under $d^2$: for any $a, b > 0$, $a^2 + b^2 < (a + b)^2$, so the two-hop cost is strictly less than the cost a direct edge of maximal admissible length $a + b$ would incur. Large behavioral transitions thus incur super-linear cost, and minimum-cost path search naturally routes through peer-grounded intermediate nodes, decomposing what would otherwise be a single large prescription into incremental, demonstrated behavioral steps (Proof 3, Appendix~\ref{app:proofs}).

\begin{algorithm}[t]
\caption{Behavioral Progression Graph Construction}
\label{alg:bpg}
\begin{algorithmic}[1]
\REQUIRE Cohort observations $\{x_i\}_{i=1}^{M}$; behavioral vectors $\{\mathbf{v}_i\}$; health outcomes $\{H(x_i)\}$; per-patient chronological ordering; clinical step sizes $\{\delta_k\}$; feature weights $\{w_k\}$
\ENSURE Behavioral Progression Graph $G = (V, E, W)$
\STATE $V \leftarrow \{x_i\}_{i=1}^{M}$;\; $E \leftarrow \emptyset$;\; $W \leftarrow \emptyset$
\FORALL{individual $p \in \{1, \ldots, P\}$}
    \STATE Sort observations of $p$ as $\mathbf{v}_p^{(1)}, \ldots, \mathbf{v}_p^{(D_p)}$ chronologically
    \STATE $\varepsilon_p \leftarrow \max_{t=1,\ldots,D_p-1} \; d\!\left(\mathbf{v}_p^{(t)},\, \mathbf{v}_p^{(t+1)}\right)$
\ENDFOR
\STATE $\varepsilon \leftarrow \min_p\;\varepsilon_p$
\FORALL{ordered pair $(x_i, x_j) \in V \times V,\; i \neq j$}
    \STATE $d_{ij} \leftarrow d(\mathbf{v}_i,\, \mathbf{v}_j)$
    \IF{$d_{ij} \leq \varepsilon$ \textbf{and} $H(x_j) > H(x_i)$}
        \STATE $E \leftarrow E \cup \{(x_i \to x_j)\}$
        \STATE $W(x_i,\, x_j) \leftarrow d_{ij}^{\;2}$
    \ENDIF
\ENDFOR
\STATE $G \leftarrow (V, E, W)$
\RETURN $G = (V, E, W)$
\end{algorithmic}
\end{algorithm}

\subsubsection{Graph Construction Algorithm}
Algorithm~\ref{alg:bpg} assembles the preceding components into the complete construction procedure. The cohort-wide threshold $\varepsilon$ is derived first, then directed edges are admitted over all ordered node pairs satisfying the joint proximity and outcome-improvement conditions of~\eqref{eq:adjacency}, with weights assigned as $d^2$ per~\eqref{eq:weight}. The graph is constructed once and supports any downstream counterfactual query without modification or retraining.

\subsection{Path Finding}
\label{sec:path_finding}

By construction, the directed paths from $x$ to $x'$ in $G$ are exactly the sequences in $\Pi(x, x')$. The path-finding function $F$ therefore searches over the directed paths of $G$. In this work, the criterion $J$ is the total edge weight along the path, so $F$ is defined as
\begin{equation}
    \pi^{*} = F(x, x') = \operatorname*{arg\,min}_{\pi \in \Pi(x, x')} \sum_{(x_i, x_j) \in \pi} W(x_i, x_j)
    \label{eq:path}
\end{equation}
where $(x_i, x_j) \in \pi$ denotes two consecutive states on $\pi$. Dijkstra's algorithm computes $F$ exactly since all weights are non-negative. Other criteria $J$, such as the number of steps, run on the same graph without modification.

For the returned path, a predictor $\hat{H}_p : \mathbb{R}^{2K+1} \rightarrow \mathbb{R}$ trained on the records of individual $p$, to whom the source $x$ belongs, estimates the outcome that individual would achieve at each step. Each step of $\pi^{*}$ is a transition between two consecutive states, so $\hat{H}_p$ is defined on transitions rather than on single states. For two consecutive states $x_i$ and $x_j$ on $\pi^{*}$, its input combines the behavioral feature vector $\mathbf{v}_i \in \mathbb{R}^K$ of the current state, the behavioral change $\mathbf{v}_j - \mathbf{v}_i \in \mathbb{R}^K$, and the current outcome $\hat{y}_i \in \mathbb{R}$, for $2K + 1$ inputs in total. It returns $\hat{y}_j = \hat{H}_p(\mathbf{v}_i,\, \mathbf{v}_j - \mathbf{v}_i,\, \hat{y}_i)$. At the source, the outcome is observed, so $\hat{y}_i = H(x)$ when $x_i = x$. The predictor takes no part in graph construction or in the selection of $\pi^{*}$. It annotates the returned path so that each step is presented with an individual-specific outcome estimate alongside the outcome observed in the cohort. Algorithm~\ref{alg:path} summarizes the procedure.

\begin{algorithm}[t]
\caption{Path Finding}
\label{alg:path}
\begin{algorithmic}[1]
\REQUIRE Behavioral Progression Graph $G = (V, E, W)$; source $x$; target $x'$; predictor $\hat{H}_p$ of the individual $p$ to whom $x$ belongs
\ENSURE Path $\pi^{*}$ and predicted outcomes $\{\hat{y}_j\}$
\STATE $\pi^{*} \leftarrow \text{Dijkstra}(G, x, x')$
\IF{no path exists}
    \RETURN $\emptyset$
\ENDIF
\STATE $\hat{y}_i \leftarrow H(x)$ for $x_i = x$
\FORALL{consecutive states $(x_i, x_j)$ on $\pi^{*}$, in path order}
    \STATE $\hat{y}_j \leftarrow \hat{H}_p(\mathbf{v}_i,\, \mathbf{v}_j - \mathbf{v}_i,\, \hat{y}_i)$
\ENDFOR
\RETURN $\pi^{*}, \{\hat{y}_j\}$
\end{algorithmic}
\end{algorithm}

\section{Experimental Setup}

POROS is domain-agnostic and applicable to any chronic disease management context. It requires only that behavioral observations can be characterized by a feature vector, MCID values~$\delta_k$, and a scalar health outcome function~$H$. To ground evaluation in a concrete and clinically demanding instantiation, we apply POROS to T1D management as a representative case, specifying these inputs accordingly.

We evaluate on two independent T1D cohorts: T1D-UOM~\cite{alsuhaymi2025}, a publicly available dataset of 15 individuals (854 patient-days), and ExActHealth, which we collected for this study, of 18 individuals (483 patient-days). Full dataset descriptions are provided in Appendix~\ref{app:datasets}. In both cohorts, each patient-day constitutes one node with behavioral feature vector $\mathbf{v}_i \in \mathbb{R}^6$ encoding six modifiable daily behaviors across dietary intake, insulin dosing, and meal timing. The $\delta_k$ configuration in Table~\ref{tab:features} is informed by prior studies and clinical guidance~\cite{arefeen2025glytwin,isaksson2024effect,peters2017,mozzillo2022optimal} and calibrated empirically so that each value corresponds to a meaningful daily change and no single feature dominates the CBTD. The health outcome $H(x_i)$ is daily TIR with clinical threshold $\tau = 70\%$ per consensus guidelines~\cite{battelino2019,ada2024}. We term nodes with $H(x_i) \geq \tau$ as BLUE (well-controlled) and those with $H(x_i) < \tau$ as RED (poorly controlled); this notation is used throughout.

The predictor $\hat{H}_p$ is instantiated as a ridge regressor trained separately for each individual on all pairs of consecutive observations in chronological order. For each pair, the input is the behavioral feature vector of the first day, the behavioral change to the second day, and the TIR of the first day. The target is the TIR of the second day. The $\ell_2$ penalty of ridge regression keeps the estimate stable when each individual contributes only a few dozen training pairs. Its held-out performance and a comparison with baselines are reported in Section~\ref{sec:results_predictor}, and the hyperparameters of all models are listed in Appendix~\ref{app:hyperparameters}.

\begin{table}[t]
\centering
\caption{Behavioral features comprising $\mathbf{v}_i$ and the $\delta_k$ configuration used in all experiments.}
\label{tab:features}
{\small
\begin{tabularx}{\columnwidth}{Xlc}
\toprule
Feature                    & Unit    & $\delta_k$ \\
\midrule
Daily carbohydrate intake  & grams   & 5          \\
Carbohydrates per meal     & grams   & 5          \\
Inter-meal interval        & minutes & 30         \\
Daily insulin dose         & units   & 1          \\
Bolus dose per meal        & units   & 1          \\
Bolus timing offset        & minutes & 10         \\
\bottomrule
\end{tabularx}}
\end{table}

\section{Results}
\label{sec:results}

\subsection{Outcome Predictor Performance}
\label{sec:results_predictor}

First, we assess how accurately $\hat{H}_p$ estimates TIR for the individual it was trained on. For each individual, pairs of consecutive observations are ordered chronologically. The first 90\% of pairs form the training set and the last 10\% form the test set, so every prediction concerns a later transition the model has not seen. This yields 415 training and 50 test pairs across 18 individuals in ExActHealth, and 748 training and 89 test pairs across 14 individuals in T1D-UOM. We compare $\hat{H}_p$ with two reference predictors. Patient Mean predicts the mean training TIR of the individual. Persistence predicts that the TIR of the next day equals that of the current day.

Table~\ref{tab:predictor} reports the results. The ridge regressor achieves the lowest error on both cohorts, with an RMSE of 9.8~pp on ExActHealth and 14.8~pp on T1D-UOM. These values are 1.2~pp and 1.1~pp below Patient Mean, and $R^2$ rises from 0.27 to 0.42 and from 0.35 to 0.44, respectively.

\begin{table*}[t]
\centering
\caption{Held-out performance of the outcome predictor and the reference models. Stochastic models are averaged over five seeds.}
\label{tab:predictor}
{\small
\begin{tabular}{lcccccc}
\toprule
& \multicolumn{3}{c}{\cellcolor{exactcol}\textbf{ExActHealth}} & \multicolumn{3}{c}{\cellcolor{uomcol}\textbf{T1D-UOM}}                         \\
\cmidrule(lr){2-4} \cmidrule(lr){5-7}
Model                                  & RMSE (pp)      & MAE (pp)      & $R^2$            & RMSE (pp)      & MAE (pp)       & $R^2$            \\
\midrule
Patient Mean                           & 10.98          & 9.25          & 0.27             & 15.95          & 12.01          & 0.35             \\
Persistence                            & 13.92          & 10.66         & $-0.17$          & 22.28          & 17.77          & $-0.27$          \\
Linear Regression                      & 13.97          & 10.87         & $-0.18$          & 15.83          & 12.44          & 0.36             \\
Random Forest                          & 10.89          & 8.86          & 0.28             & 15.66          & 12.36          & 0.37             \\
XGBoost                                & 11.43          & 8.94          & 0.21             & 16.93          & 13.09          & 0.27             \\
MLP                                    & 26.61          & 20.47         & $-3.28$          & 25.77          & 20.22          & $-0.70$          \\
\textbf{Ridge Regressor ($\hat{H}_p$)} & \textbf{9.77}  & \textbf{8.16} & \textbf{0.42}    & \textbf{14.84} & \textbf{11.85} & \textbf{0.44}    \\
\bottomrule
\end{tabular}}
\end{table*}

\subsection{Graph Properties}
\label{sec:graph_properties}

\begin{figure}[t]
\centering
\includegraphics[width=0.95\linewidth]{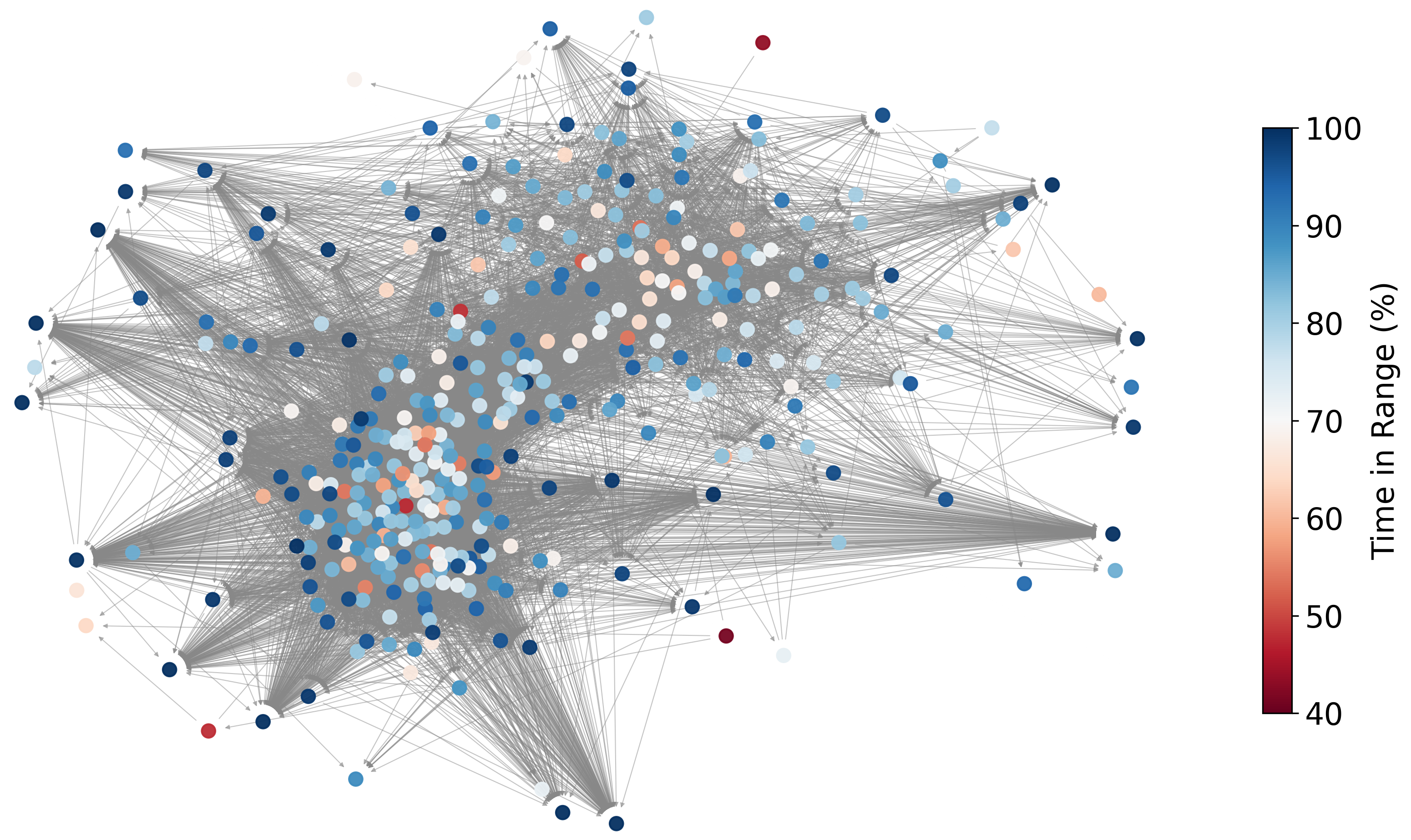}
\caption{Behavioral Progression Graph over 483 patient-days and 25,556 directed edges (ExActHealth). Node color indicates TIR value from red (low) to blue (high).}
\label{fig:graph}
\end{figure}

After applying POROS independently to each cohort, we construct the corresponding Behavioral Progression Graphs using uniform feature weights ($w_k = 1$). The resulting thresholds are $\varepsilon = 10.95$ for ExActHealth and $\varepsilon = 14.72$ for T1D-UOM. Figure~\ref{fig:graph} visualizes the ExActHealth graph, where nodes are colored according to TIR using a red-to-blue gradient; edges represent peer-grounded behavioral transitions that satisfy both the CBTD constraint and strict TIR improvement condition.

Table~\ref{tab:graph} summarizes the structural properties of the two resulting graphs. The two cohorts exhibit substantially different graph characteristics. In particular, RED nodes account for 20.1\% of ExActHealth but 45.4\% of T1D-UOM, and the larger cohort admits nearly $4.1\times$ more directed edges despite having only $1.8\times$ more nodes. Mean out-degree increases substantially in T1D-UOM (122.3 vs. 52.9), indicating a denser behavioral neighborhood structure. Because no individual identity constraint appears in Equation~\ref{eq:adjacency}, edges form freely across patient boundaries. As reported in Table~\ref{tab:graph}, 98.1\% and 97.0\% of multi-hop paths in ExActHealth and T1D-UOM, respectively, traverse patient boundaries, structurally embedding social comparison into every recommendation POROS produces.

\begin{table}[t]
\centering
\caption{Cohort and graph summary.}
\label{tab:graph}
{\small
\begin{tabularx}{\columnwidth}{Xll}
\toprule
& \cellcolor{exactcol}\textbf{ExActHealth} & \cellcolor{uomcol}\textbf{T1D-UOM} \\
\midrule
Subjects            & 18               & 15            \\
Days                & 483              & 854           \\
RED nodes           & 97~(20.1\%)      & 388~(45.4\%)  \\
BLUE nodes          & 386~(79.9\%)     & 466~(54.6\%)  \\
Directed edges      & 25{,}556         & 104{,}404     \\
Mean out-degree     & 52.9             & 122.3         \\
Cross-patient paths & 98.1\%           & 97.0\%        \\
$\varepsilon$       & 10.95            & 14.72         \\
\bottomrule
\end{tabularx}}
\end{table}

\subsection{Counterfactual Path Quality}
\label{sec:results_quality}

For any source node, POROS finds the minimum-cost path to every reachable target with strictly higher TIR, decomposing what would otherwise be a single behavioral jump into incremental peer-grounded steps. To quantify this decomposition, we compare the TIR gain $\hat{H}_p$ predicts for the source individual after a single direct transition with the average predicted TIR gain at each step of the corresponding multi-hop path. We evaluate two source populations in each cohort:

\begin{itemize}
    \item \textbf{All}: every source node in the graph.
    \item \textbf{RED}: source nodes with TIR~$<70\%$ --- the primary clinical target population.
\end{itemize}

Table~\ref{tab:tir_quality} summarizes the results. POROS consistently reduces the predicted outcome improvement at each step. In ExActHealth, the average direct-path gain decreases from $3.4 \pm 24.4$~pp to $0.3 \pm 10.0$~pp per multi-hop step (10.0$\times$ reduction), while for RED sources it decreases from $12.1 \pm 25.2$~pp to $2.0 \pm 10.6$~pp per multi-hop step (6.1$\times$ reduction). A similar pattern is observed in T1D-UOM despite its larger and more heterogeneous population: the average direct-path gain decreases from $4.0 \pm 26.2$~pp to $1.1 \pm 15.4$~pp per multi-hop step across \textbf{All} sources (3.6$\times$ reduction), and from $9.4 \pm 24.5$~pp to $2.0 \pm 14.7$~pp per multi-hop step for \textbf{RED} sources (4.8$\times$ reduction). These results show that POROS consistently transforms large behavioral jumps into substantially smaller, incremental improvements across both cohorts, and RED sources are predicted to gain more per step than All sources on both cohorts.

\begin{table}[tp]
\centering
\caption{Comparison of direct transitions and POROS multi-hop paths. \textbf{All} includes every source node, whereas \textbf{RED} includes only source nodes with TIR~$<70\%$. Direct is the TIR gain $\hat{H}_p$ predicts for the source individual after a single jump to the target; MH/step is the mean predicted TIR gain per step along the minimum-cost path. Reduction~$=$~mean Direct~$\div$~mean MH/step.}
\label{tab:tir_quality}
{\small
\begin{tabularx}{\columnwidth}{X c c c c}
\toprule
    & $n$ pairs   & Direct (pp)      & MH/step (pp)     & Reduction              \\
\midrule
\multicolumn{5}{c}{\cellcolor{exactcol}\textbf{ExActHealth}}                     \\
All & 111{,}289   & $3.4 \pm 24.4$   & $0.3 \pm 10.0$   & \textbf{10.0$\times$}  \\
RED &  37{,}442   & $12.1 \pm 25.2$  & $2.0 \pm 10.6$   & \textbf{6.1$\times$}   \\
\midrule
\multicolumn{5}{c}{\cellcolor{uomcol}\textbf{T1D-UOM}}                           \\
All & 288{,}887   & $4.0 \pm 26.2$   & $1.1 \pm 15.4$   & \textbf{3.6$\times$}   \\
RED & 180{,}808   & $9.4 \pm 24.5$   & $2.0 \pm 14.7$   & \textbf{4.8$\times$}   \\
\bottomrule
\end{tabularx}}
\end{table}

\subsection{Comparison with Baselines}
\subsubsection{Endpoint-only Baselines}
\label{sec:results_baseline}

\begin{table*}[t]
\centering
\caption{An endpoint-only baseline inspired by Wachter et al. and DiCE versus POROS on the RED~$\rightarrow$~nearest-BLUE setting. Each RED source is classified by whether the direct jump is feasible ($d \leq \varepsilon$) and whether POROS resolves it as a single- or multi-hop path. $n$ is the number of RED source nodes per cohort; cells give the count and share of sources in each case. POROS finds a feasible path for 97.9\% (ExActHealth) and 95.6\% (T1D-UOM) of RED sources.}
\label{tab:baseline}
{\small
\begin{tabularx}{\textwidth}{@{}c c l l X@{}}
\toprule
Case & Condition & \cellcolor{exactcol}\textbf{ExActHealth}~($n{=}97$) & \cellcolor{uomcol}\textbf{T1D-UOM}~($n{=}388$) & Interpretation   \\
\midrule
1    & $d \leq \varepsilon$  & 81 (83.5\%)     & 294 (75.7\%)    & POROS single-hop --- both methods agree                \\
2    & $d \leq \varepsilon$  & 10 (10.3\%)     & 53 (13.6\%)     & POROS multi-hop --- $d^2$ prefers gradual steps        \\
3    & $d > \varepsilon$     & 4 (4.1\%)       & 24 (6.2\%)      & Endpoint-only infeasible --- POROS routes via peers    \\
4    & $d > \varepsilon$     & 2 (2.1\%)       & 17 (4.4\%)      & Both methods fail --- target behaviorally isolated     \\
\bottomrule
\end{tabularx}}
\end{table*}

Endpoint-only methods such as Wachter et al.~\cite{wachter2017} and DiCE~\cite{mothilal2020} return a single well-controlled state and expect the patient to close the entire behavioral gap in one move. We therefore construct an endpoint-only baseline that follows the minimal-distance formulation of these methods, nominating the observed BLUE node nearest to the source as its single target. To compare on equal footing, we give the baseline and POROS an identical task: every RED node is a source, and its target is its nearest BLUE node by raw distance $d$. The two then differ only in what they return: the direct jump, versus the minimum-cost POROS path to that target through the $\varepsilon$-graph. We classify each source into four cases by whether the jump is feasible ($d \leq \varepsilon$) and whether POROS resolves it as a single or multi-hop path (Table~\ref{tab:baseline}).

\begin{figure}[t]
\centering
\includegraphics[width=0.83\linewidth]{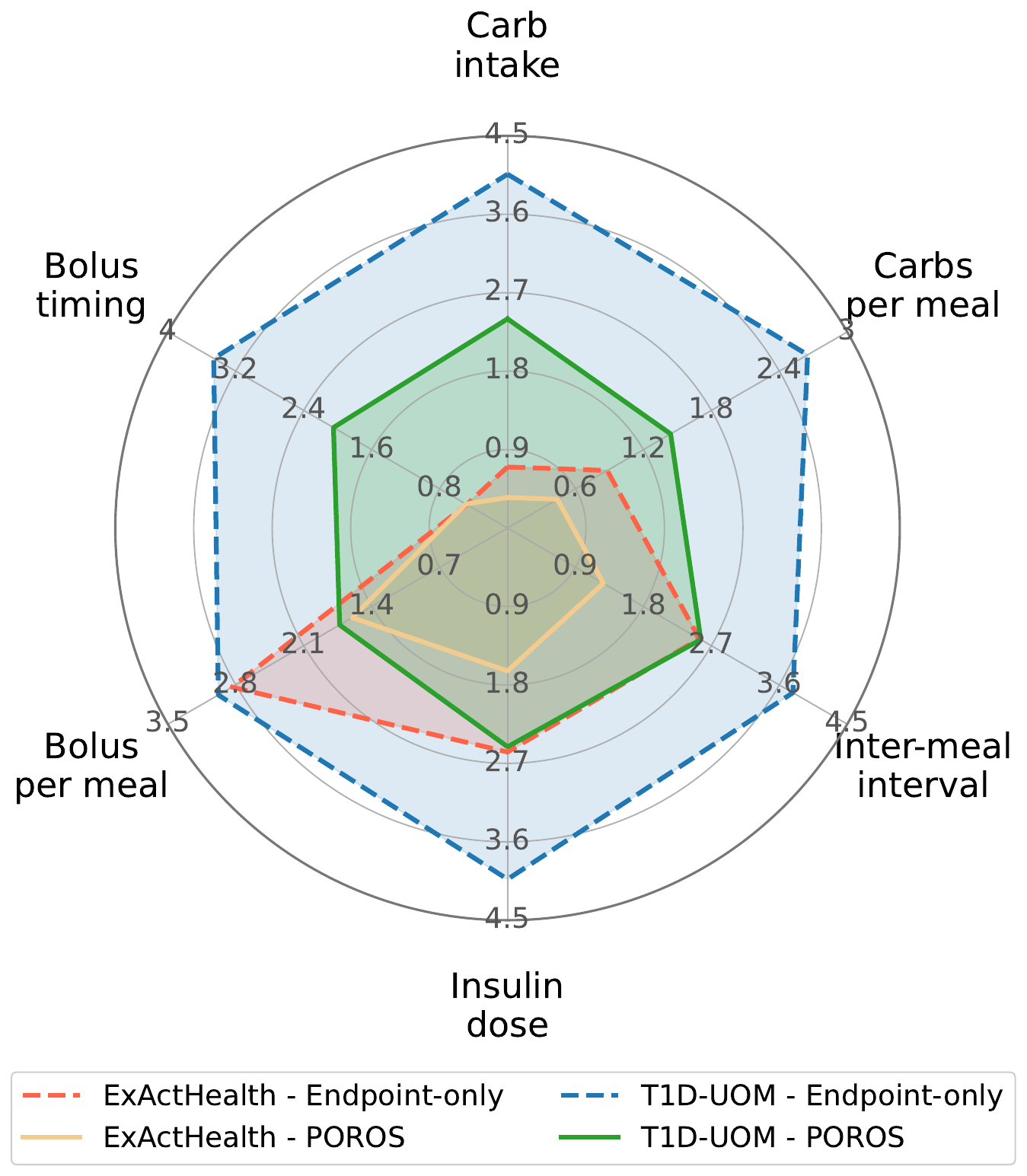}
\caption{Per-feature behavioral change (MCID units) for the RED sources where POROS returns a multi-hop path while endpoint-only methods prescribe a single jump. Dashed contours denote the endpoint-only single jump; solid contours denote the mean POROS change per step. On both cohorts POROS reduces the change demanded of every behavior.}
\label{fig:baseline_radar}
\end{figure}

In most cases the target is one feasible hop away and POROS returns exactly that hop, coinciding with the endpoint-only recommendation for 81 (83.5\%) of ExActHealth and 294 (75.7\%) of T1D-UOM RED sources (Case 1); it does not manufacture intermediate steps when a direct transition is already actionable. POROS instead returns a multi-hop path for 14.4\% and 19.8\% of sources respectively---cases where the baseline nominates the same target but prescribes closing the whole gap at once, most common in the larger, more heterogeneous T1D-UOM cohort. Within these, the direct jump exceeds $\varepsilon$ for 4.1\% and 6.2\% of sources (Case 3), where POROS recovers a feasible route through peers while endpoint-only methods return a prescription that exceeds the cohort-derived bound and therefore has no direct edge in the graph. For the remaining 2.1\% and 4.4\% (Case 4), the target is behaviorally isolated and POROS returns no path at all, whereas an endpoint-only method still emits a target regardless of whether the change it demands is attainable.

By spreading the adjustment across a median of two hops, POROS roughly halves the per-step behavioral change---reductions of 41\% in carbohydrate intake, 38\% in insulin dose, and 42\% in bolus timing on T1D-UOM---so each step is a modest, attainable modification rather than a wholesale overhaul, directly serving self-efficacy while reaching the same destination (Figure~\ref{fig:baseline_radar}).

\subsubsection{Path-aware Baselines}
\label{sec:results_face}

Path-aware methods construct a graph over observed instances and route through high-density regions toward a counterfactual endpoint. We instantiate FACE~\cite{poyiadzi2020} on the same patient-day nodes under the same distance $d$, so that only the routing mechanism differs. Its graph connects each node to its $k = 10$ nearest neighbors without direction. Every edge is weighted by the kernel density at the midpoint scaled by $d$.

\begin{table}[t]
\centering
\caption{FACE versus POROS on identical source--target pairs. Sources are RED nodes and targets are BLUE nodes with strictly higher TIR. FACE routes over an undirected $k$-NN graph ($k = 10$) with density-weighted edges, while POROS routes over the $\varepsilon$-graph with $d^2$ weights. TIR along each path is predicted by $\hat{H}_p$ for the source individual. $n$ is the number of pairs both methods reach.}
\label{tab:face}
{\small
\begin{tabularx}{\columnwidth}{X c c}
\toprule
                                  & FACE             & POROS              \\
\midrule
\multicolumn{3}{c}{\cellcolor{exactcol}\textbf{ExActHealth}~($n{=}33{,}632$)} \\
Infeasible paths$^{\dagger}$ (\%)       & 3.5              & 0.0                \\
Monotone paths$^{\ddagger}$ (\%)   & 14.2             & 19.1               \\
Hops per path                     & $6.48 \pm 3.77$  & $4.90 \pm 2.43$    \\
TIR gain per step (pp)            & $1.5 \pm 8.7$    & $2.1 \pm 10.7$     \\
\midrule
\multicolumn{3}{c}{\cellcolor{uomcol}\textbf{T1D-UOM}~($n{=}162{,}660$)}      \\
Infeasible paths$^{\dagger}$ (\%)       & 3.4              & 0.0                \\
Monotone paths$^{\ddagger}$ (\%)   & 12.3             & 14.9               \\
Hops per path                     & $6.97 \pm 3.55$  & $5.57 \pm 2.65$    \\
TIR gain per step (pp)            & $1.6 \pm 10.8$   & $2.0 \pm 14.8$     \\
\bottomrule
\end{tabularx}}
{\footnotesize\raggedright $^{\dagger}$Contain at least one step with $d > \varepsilon$. $^{\ddagger}$Predicted TIR improves at every step.\par}
\end{table}

Table~\ref{tab:face} summarizes the comparison. Under the TIR predicted by $\hat{H}_p$ for the source individual, POROS produces more monotone paths than FACE on both cohorts. Its paths improve the predicted TIR at every step for 19.1\% of pairs on ExActHealth and 14.9\% on T1D-UOM, against 14.2\% and 12.3\% for FACE. POROS paths contain 4.90 and 5.57 hops against 6.48 and 6.97 for FACE, with a predicted TIR gain of 2.1~pp and 2.0~pp per step against 1.5~pp and 1.6~pp. FACE also breaches the feasibility threshold POROS enforces. On ExActHealth 3.5\% of its paths contain a step above $\varepsilon$ and on T1D-UOM 3.4\% do (0.5\% of all steps on both cohorts), a single-period behavioral change that exceeds the cohort-derived bound.

\subsubsection{Within-Class Transitions}
\label{sec:results_within}

Classifier-based CE methods define a counterfactual by flipping a classifier's decision. Wachter et al.~\cite{wachter2017} and DiCE~\cite{mothilal2020} search for the nearest point on the other side of the decision boundary, and FACE~\cite{poyiadzi2020} routes toward a target that a classifier assigns to the desired class. These methods therefore define no target when the source and the desired state fall in the same class. In T1D, for instance, a patient whose TIR rises from 55\% to 65\% remains poorly controlled, and a patient whose TIR rises from 75\% to 85\% remains well controlled. Both are clinically meaningful improvements, yet neither crosses the 70\% threshold. POROS imposes no such restriction, because an edge requires only that the outcome strictly improves. The same graph therefore answers within-class queries without modification.

We evaluate two such queries in each cohort. BLUE~$\rightarrow$~BLUE takes every BLUE node as a source and every BLUE node with strictly higher TIR as a target. RED~$\rightarrow$~RED does the same within the RED nodes. Table~\ref{tab:within} reports the results. POROS reaches 78.1\% of BLUE~$\rightarrow$~BLUE pairs on ExActHealth and 83.7\% on T1D-UOM, with a mean of 3.94 and 4.05 hops per path. For RED~$\rightarrow$~RED, it reaches 46.3\% and 67.6\% of pairs with 2.55 and 3.87 hops. Reach is lower for RED~$\rightarrow$~RED, most clearly on ExActHealth, where RED nodes make up only 20.1\% of the graph. These results show that POROS supports improvement between any two outcome-ordered states, including transitions that classifier-based methods cannot express.

\begin{table}[t]
\centering
\caption{Within-class queries answered by POROS. Sources and targets belong to the same class, and each target has strictly higher TIR than its source. Reached is the number of pairs POROS connects through the $\varepsilon$-graph.}
\label{tab:within}
{\small
\begin{tabularx}{\columnwidth}{X c c}
\toprule
                  & \cellcolor{exactcol}\textbf{ExActHealth} & \cellcolor{uomcol}\textbf{T1D-UOM} \\
\midrule
\multicolumn{3}{c}{\textit{BLUE~$\rightarrow$~BLUE}}                                              \\
\midrule
Pairs             & 73{,}847                     & 108{,}079                    \\
Reached           & 57{,}671~(78.1\%)            & 90{,}450~(83.7\%)            \\
Hops per path     & $3.94 \pm 2.07$              & $4.05 \pm 2.22$              \\
\midrule
\multicolumn{3}{c}{\textit{RED~$\rightarrow$~RED}}                                                \\
\midrule
Pairs             & 4{,}620                      & 75{,}022                     \\
Reached           & 2{,}137~(46.3\%)             & 50{,}698~(67.6\%)            \\
Hops per path     & $2.55 \pm 1.41$              & $3.87 \pm 2.03$              \\
\bottomrule
\end{tabularx}}
\end{table}

\subsection{Sensitivity and Ablation}
\subsubsection{Sensitivity to $\varepsilon$}
\label{sec:results_epsilon}

\begin{figure}[t]
\centering
\subfloat[ExActHealth]{%
    \includegraphics[width=\columnwidth]{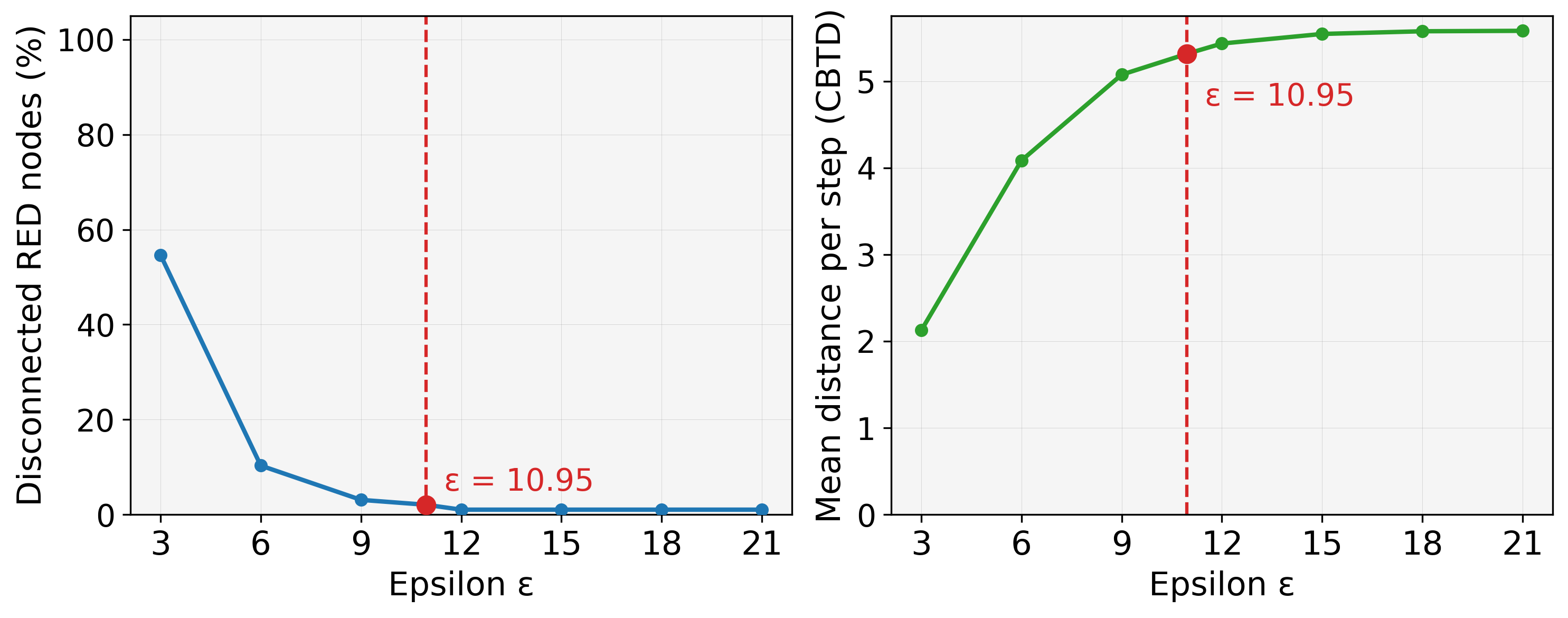}%
    \label{fig:epsilon_exact}}\\
\subfloat[T1D-UOM]{%
    \includegraphics[width=\columnwidth]{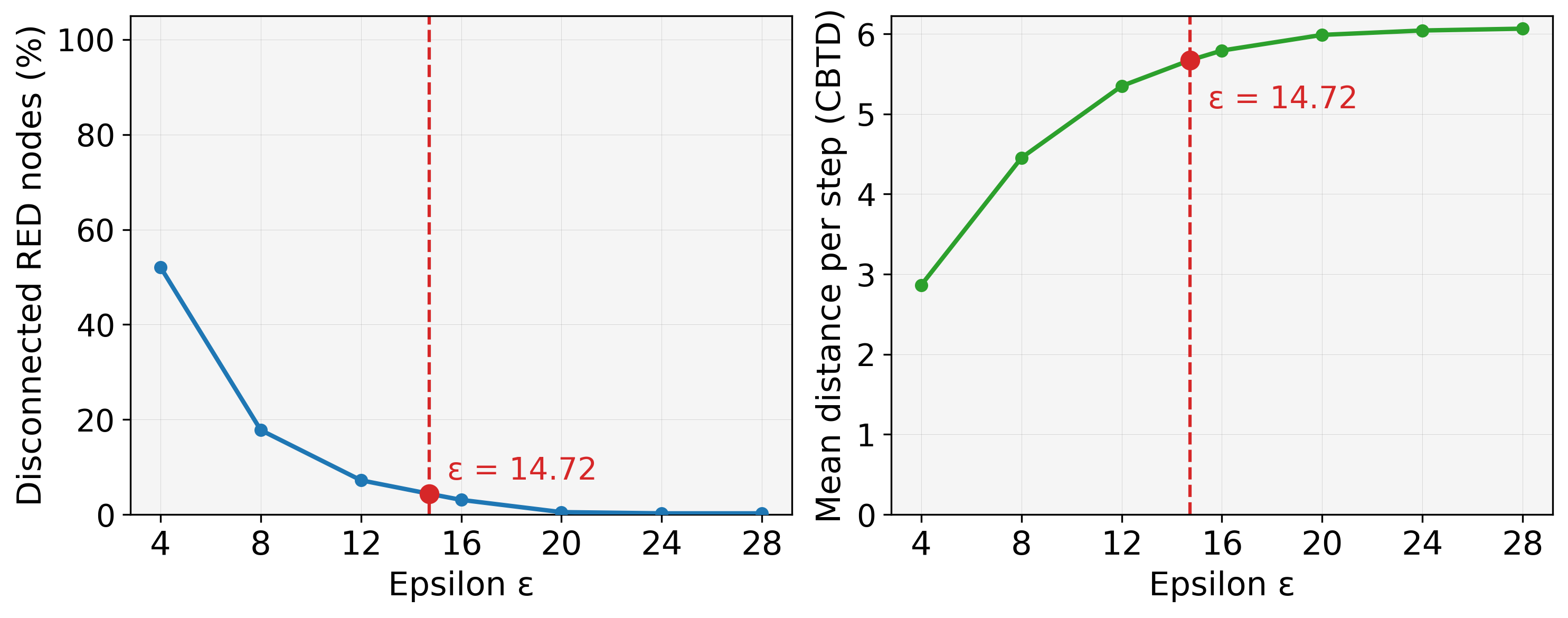}%
    \label{fig:epsilon_uom}}
\caption{Sensitivity of graph connectivity and step size to $\varepsilon$ for (a) ExActHealth and (b) T1D-UOM. In each row, the left panel shows the percentage of RED sources with no path to any BLUE target, and the right panel shows the mean CBTD per step across all paths from RED sources. The red dashed line marks the derived $\varepsilon$.}
\label{fig:epsilon_sensitivity}
\end{figure}

The threshold $\varepsilon$ is the principal structural parameter of the Behavioral Progression Graph, governing which behavioral transitions are admitted as edges. Rather than treating it as a free hyperparameter, we derive $\varepsilon$ directly from each cohort's observed within-subject behavioral dynamics, yielding $\varepsilon = 10.95$ for ExActHealth and $\varepsilon = 14.72$ for T1D-UOM, each limited by the most behaviorally constrained individual in that cohort. Figure~\ref{fig:epsilon_sensitivity} examines the consequences of departing from this value in either direction, for both datasets.

Too tight an $\varepsilon$ leaves poorly controlled sources with no route to any well-controlled target: 54.6\% of RED nodes are disconnected at $\varepsilon = 3$ in ExActHealth and 52.1\% at $\varepsilon = 4$ in T1D-UOM. Disconnection falls steeply as $\varepsilon$ grows and reaches 2.1\% and 4.4\% at the derived thresholds, with a mean step of 5.31 and 5.67 CBTD units. Widening $\varepsilon$ further gains little connectivity, reaching 1.0\% at $\varepsilon = 21$ and 0.3\% at $\varepsilon = 28$, while admitting steps larger than every individual in the cohort has demonstrated. The derived $\varepsilon$ therefore sits at the point where the graph is nearly fully connected within demonstrated behavioral capacity, without manual tuning.

\subsubsection{Edge Weighting: $d$ vs $d^2$}
\label{sec:results_d2}

The $d^2$ edge weight is motivated through the triangle inequality: under a linear weight $d$, a direct edge is never more costly than the sum of its parts, so Dijkstra always prefers the largest admissible single hop wherever one exists, collapsing paths to the behavioral jumps that self-efficacy theory identifies as inactionable. The quadratic weight breaks this property in cost space, penalizing large single steps super-linearly and routing the search through peer-grounded intermediate nodes when they exist.

\begin{figure}[t]
\centering
\subfloat[ExActHealth]{%
    \includegraphics[width=0.48\columnwidth]{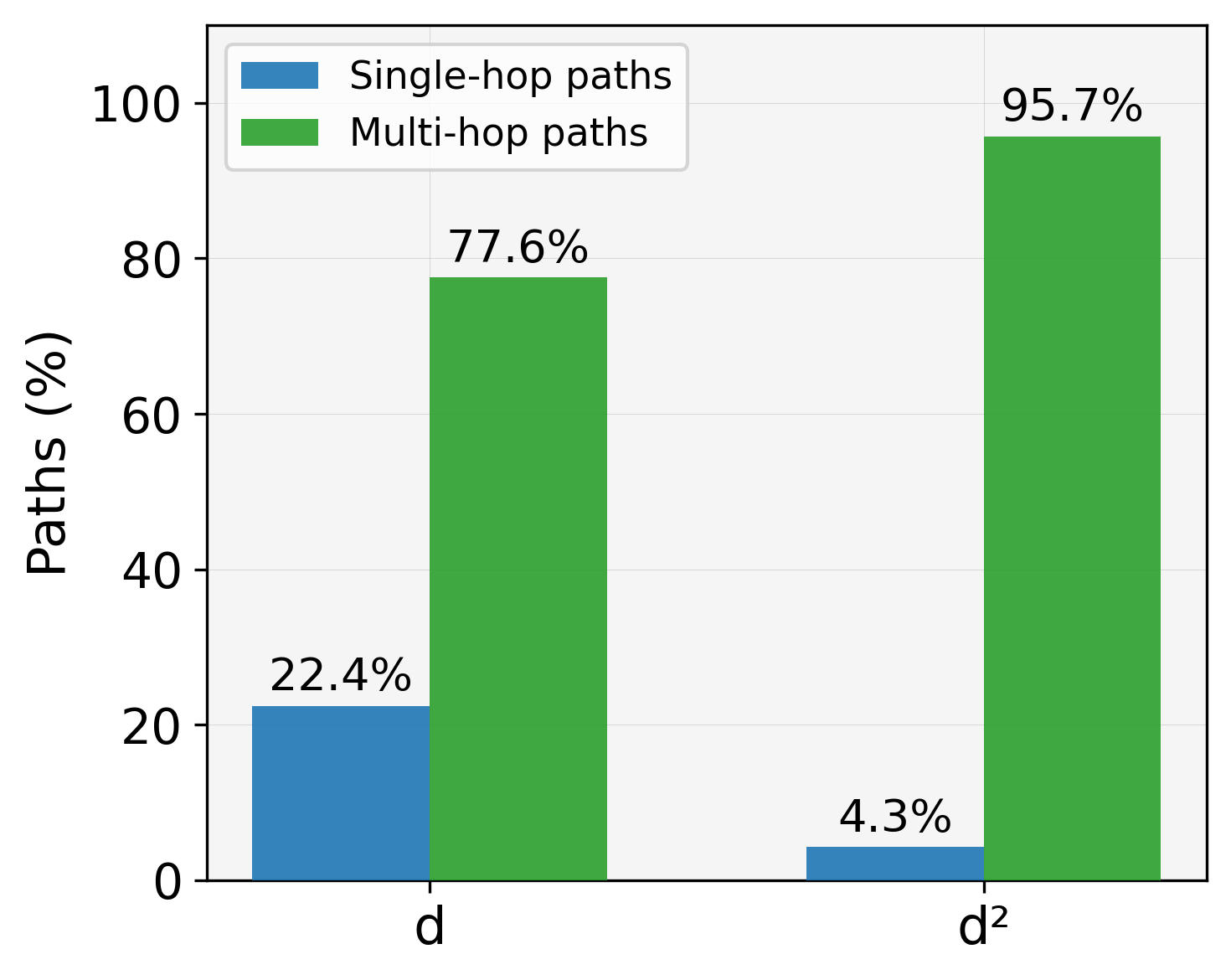}%
    \label{fig:d2_exact}}
\hfill
\subfloat[T1D-UOM]{%
    \includegraphics[width=0.48\columnwidth]{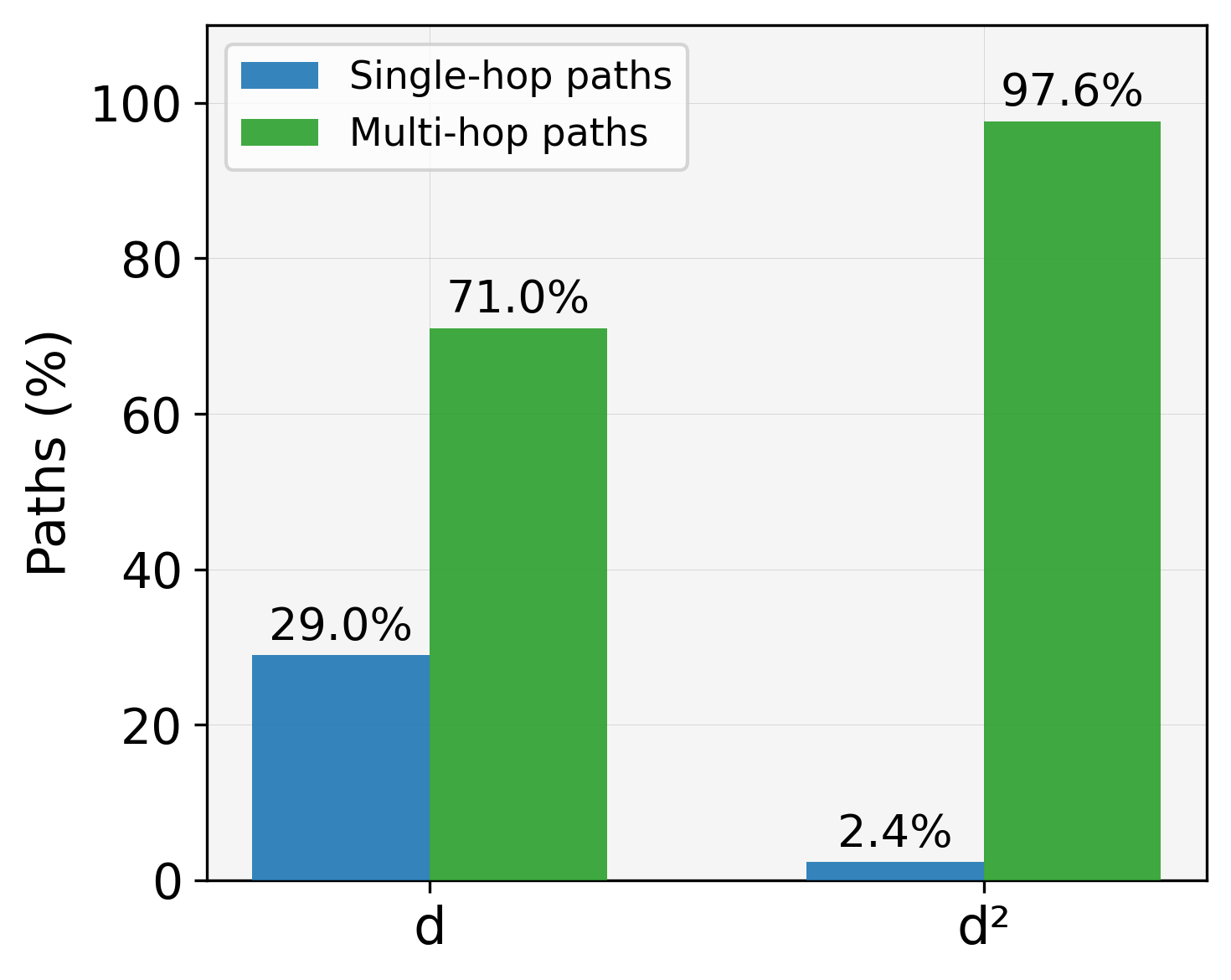}%
    \label{fig:d2_uom}}
\caption{Effect of edge weighting on path structure for (a) ExActHealth and (b) T1D-UOM. Under linear $d$, Dijkstra favors large single-hop transitions, while under $d^2$ the search is routed through peer-grounded intermediate nodes.}
\label{fig:d_vs_d2}
\end{figure}

Figure~\ref{fig:d_vs_d2} confirms this empirically across both datasets. Under linear $d$ weighting, 29.0\% of paths from poorly-controlled source nodes to well-controlled targets resolve as single-hop direct jumps in T1D-UOM; under $d^2$, this falls to 2.4\%, a 91.9\% reduction, with 97.6\% of paths now traversing multiple peer-grounded intermediate steps. The same pattern holds in T1D-Anon, where single-hop paths fall from 22.4\% under $d$ to 4.3\% under $d^2$, an 80.8\% reduction. This shift occurs without any change to graph structure or the set of admitted edges in either cohort: the same principled-$\varepsilon$ graph, simply reweighted, produces substantially more incremental trajectories by construction.

\section{Discussion}
\label{sec:discussion}

POROS carries some limitations worth acknowledging. First, because it is observational and built on peer-grounded data, a path does not establish that the source patient's outcome would rise in the same way after adopting those behaviors. The predictor $\hat{H}_p$ therefore plays an important role by estimating the outcome the source patient would actually obtain after changing their behavior. However, factors outside the behavioral feature vector, such as day-to-day variation in insulin sensitivity, glycemic carryover, stress, illness, sleep, and physical activity, shape the glycemic response to the same behavioral step. These factors limit what any behavioral predictor can explain.

Second, the steps of a path come from peer behavior and may fall outside the behavioral range of the source individual. For instance, a patient who has never recorded a well-controlled day offers $\hat{H}_p$ no training data near a well-controlled target, so the predictor must extrapolate beyond the records it was trained on. This distribution shift affects any counterfactual method that estimates the outcome of an unobserved behavior, and resolving it requires a dedicated investigation beyond the scope of this work.

Third, path quality is bounded by the diversity of the observed cohort. Sparse or homogeneous data limits graph navigability regardless of framework design, and for a small fraction of sources no peer-grounded route to a well-controlled state exists at all.

Future work should evaluate POROS in a prospective real-world study. Such a study would measure adherence, showing whether patients presented with a peer-grounded path attempt more of it than a single prescription. The data it collects would also enlarge the records of each individual and improve the accuracy of $\hat{H}_p$. In addition, the MCID values $\delta_k$ and the feature weights $w_k$ used in this study were set for experimental evaluation. In real-world use, these hyperparameters should be adapted to each patient. Finally, extending POROS to other chronic disease settings where behavioral observations, MCID values, and a scalar health outcome are defined --- weight management, for instance --- is an open and natural direction for future work.

\section{Conclusion}
\label{sec:conclusion}

We introduced POROS, a framework that recasts counterfactual explanation for behavioral health management as path planning. POROS constructs a Behavioral Progression Graph over observed patient-day states derived from multimodal behavioral records, in which every edge requires peer-grounded behavioral proximity and strict improvement in the observed health outcome. Its Clinical Behavioral Transition Distance places heterogeneous behaviors on a common scale, and a cohort-derived threshold bounds the magnitude of each transition. Once the graph is constructed, path-finding algorithms solve the optimization problem by identifying minimum-cost paths through it, decomposing behavioral gaps into incremental steps. These steps draw on behavioral profiles that real patients have demonstrated, connecting gradual change with peer-grounded examples and embedding self-efficacy and social comparison into the recommendation. The same graph supports queries without retraining, covering transitions both across and within outcome classes.

We evaluated POROS using meal logs, insulin delivery records, and CGM-derived outcomes from two independent T1D cohorts, ExActHealth and T1D-UOM. Compared with an endpoint-only baseline, POROS returned multi-hop paths to the same target for 14.4\% and 19.8\% of source days with TIR below 70\%, respectively. Among these multi-hop cases on T1D-UOM, the mean per-step changes in carbohydrate intake, insulin dose, and bolus timing were reduced by 41\%, 38\%, and 42\%. POROS also connected 78.1\% and 83.7\% of outcome-ordered pairs of well-controlled days, demonstrating improvement paths that do not require crossing a clinical threshold. A separate per-patient predictor annotated paths with individualized TIR estimates. Compared with FACE, POROS produced shorter paths and more paths with monotonically increasing predicted TIR. Together, these results establish POROS as an observational framework that translates multimodal health records into incremental, peer-grounded paths for long-term chronic disease management, with each step satisfying the cohort-derived bound on behavioral change and strictly increasing the observed health outcome by construction.

\section*{Appendix}
\setcounter{subsection}{0}
\renewcommand{\thesubsection}{\Alph{subsection}}
\setcounter{table}{0}
\renewcommand{\thetable}{A.\Roman{table}}

\subsection{Dataset Descriptions}
\label{app:datasets}

\subsubsection*{T1D-UOM}

T1D-UOM is a publicly available longitudinal multimodal type 1 diabetes (T1D) dataset\footnote{\url{https://zenodo.org/records/15169264}} comprising 17 adults with T1D followed over an approximately three-month period under free-living conditions. Participants used continuous glucose monitors (CGM) together with insulin pumps or multiple daily injections, and the released data span glucose, insulin, nutrition, activity, and sleep. From these records we derive one patient-day node per day, described by the six behavioral features in Table~\ref{tab:features} (dietary intake, insulin dosing, and meal timing), with daily time-in-range (TIR) as the health outcome. After keeping only days with complete behavioral features, our evaluation uses 15 individuals and 854 patient-days.

\subsubsection*{ExActHealth}

ExActHealth is a prospective observational cohort we collected for this study, comprising 19 adults with T1D followed over an approximately one-month period. The study was approved by the Institutional Review Board of Arizona State University and Mayo Clinic Arizona (IRB \#24-010332), and all participants provided informed consent prior to enrollment. Participants were recruited from an outpatient endocrinology clinic, and all used the same automated insulin delivery (AID) system, a Tandem t:slim X2 insulin pump with a Dexcom CGM. They were followed under free-living conditions without any change to their standard clinical care. Each participant wore a consumer smartwatch throughout the study and used a custom mobile application to self-report daily food intake, including meal time, macronutrient content, and optional food images. Insulin pump, CGM, smartwatch, and food-log data were collected over the study period.

\subsection{Proofs}
\label{app:proofs}

\subsubsection*{Proof 1 --- Acyclicity of the Behavioral Progression Graph}
\begin{claim}
$G = (V, E, W)$ is a directed acyclic graph (DAG).
\end{claim}
\begin{proof}
Suppose for contradiction that $G$ contains a directed cycle $x_{i_1} \to x_{i_2} \to \cdots \to x_{i_m} \to x_{i_1}$. By the adjacency condition, every edge in this cycle requires a strictly increasing health outcome:
\begin{equation*}
    H(x_{i_1}) < H(x_{i_2}) < \cdots < H(x_{i_m}) < H(x_{i_1})
\end{equation*}
This requires $H(x_{i_1}) < H(x_{i_1})$, a contradiction. Therefore no directed cycle exists in $G$, and $G$ is a DAG.
\end{proof}

\subsubsection*{Proof 2 --- Peer-Groundedness Guarantee}
\begin{claim}
Every admitted edge $(x_i, x_j) \in E$ represents a behavioral transition whose magnitude does not exceed the demonstrated single-period behavioral capacity of every individual in the cohort.
\end{claim}
\begin{proof}
For individual $p$, define the maximum observed single-period behavioral transition:
\begin{equation*}
    \varepsilon_p = \max_{t=1,\ldots,D_p - 1} \; d\!\left(\mathbf{v}_p^{(t)},\, \mathbf{v}_p^{(t+1)}\right)
\end{equation*}
The global threshold is set as $\varepsilon = \min_p \; \varepsilon_p$. By construction, $\varepsilon \leq \varepsilon_p$ for every $p \in \{1, \ldots, P\}$. For any admitted edge $(x_i, x_j) \in E$, the adjacency condition requires $d(\mathbf{v}_i, \mathbf{v}_j) \leq \varepsilon$. It follows that:
\begin{equation*}
    d(\mathbf{v}_i, \mathbf{v}_j) \leq \varepsilon \leq \varepsilon_p \quad \forall\, p \in \{1, \ldots, P\}
\end{equation*}
Every admitted edge therefore represents a behavioral transition within the maximum single-period change that every individual in the cohort has demonstrably made.
\end{proof}

\subsubsection*{Proof 3 --- Cost Advantage of Multi-hop Paths Under $d^2$}
\begin{claim}
Under quadratic edge weights $W(x_i, x_j) = d(\mathbf{v}_i, \mathbf{v}_j)^2$, a two-hop path through an intermediate node $x_k$ is strictly cheaper than the direct edge $(x_i, x_j)$ whenever $d(\mathbf{v}_i, \mathbf{v}_j) > \sqrt{d(\mathbf{v}_i, \mathbf{v}_k)^2 + d(\mathbf{v}_k, \mathbf{v}_j)^2}$.
\end{claim}
\begin{proof}
Let $a = d(\mathbf{v}_i, \mathbf{v}_k)$, $b = d(\mathbf{v}_k, \mathbf{v}_j)$, and $c = d(\mathbf{v}_i, \mathbf{v}_j)$ for some $x_i, x_k, x_j \in V$ with $a, b > 0$. Since CBTD is a metric, the triangle inequality holds:
\begin{equation*}
    c \leq a + b \tag{i}
\end{equation*}
The two-hop cost is strictly less than the direct cost under $d^2$ if and only if:
\begin{equation*}
    a^2 + b^2 < c^2 \iff c > \sqrt{a^2 + b^2} \tag{ii}
\end{equation*}
The interval $\left(\sqrt{a^2 + b^2},\ a + b\right]$ is non-empty for all $a, b > 0$, since:
\begin{equation*}
    \sqrt{a^2 + b^2} < a + b \iff a^2 + b^2 < a^2 + 2ab + b^2 \iff 2ab > 0 \tag{iii}
\end{equation*}
which holds strictly for any $a, b > 0$. Therefore any $c$ in this interval simultaneously satisfies~(i) and~(ii): the direct edge exists and is geometrically reachable, yet the two-hop path is strictly cheaper under $d^2$. Concretely, for $a = b = r$ and $c = 2r$:
\begin{equation*}
    a^2 + b^2 = 2r^2 \qquad c^2 = 4r^2
\end{equation*}
the two-hop path costs exactly half the direct path. More generally:
\begin{equation*}
    \frac{a^2 + b^2}{c^2} < 1 \quad \text{for all } c > \sqrt{a^2 + b^2}
\end{equation*}
confirming that minimum-cost path search under $d^2$ strictly favors paths with smaller, distributed steps over large single transitions whenever intermediate nodes satisfying this condition are available.
\end{proof}

\subsection{Time Complexity}
\label{app:complexity}

Let $M$ be the total number of patient-day observations, $K$ the number of behavioral features, $P$ the number of individuals, and $E$ the number of admitted edges.

\paragraph{Threshold derivation.} For each individual, observations are sorted chronologically and consecutive-period CBTD values are computed. Sorting across all individuals costs $O(M \log M)$, and the $M - P$ consecutive pairs each require an $O(K)$ distance evaluation, giving $O(M \log M + MK)$ for this stage.

\paragraph{Graph construction.} The algorithm iterates over all $M(M-1)$ ordered node pairs, evaluating one CBTD ($O(K)$) and one outcome comparison ($O(1)$) per pair. Construction is therefore $O(M^2 K)$ in time and $O(M^2)$ in worst-case space for the edge set. This stage dominates the overall cost and is performed once per cohort.

\paragraph{DAG structure.} Because every edge enforces strict outcome improvement, $G$ is acyclic (Proof 1). Dijkstra's algorithm with a binary heap computes single-source shortest paths in $O((M + E) \log M)$. Single-source shortest paths on a DAG can also be computed in $O(M + E)$ by relaxing edges in topological order. We use Dijkstra for generality, but note that acyclicity permits linear-time-per-source shortest paths as cohorts scale.

\subsection{Hyperparameters}
\label{app:hyperparameters}

All regressors are trained separately for each individual, with input features standardized on that individual's training pairs. Table~\ref{tab:hyperparameters} lists their hyperparameters together with those of FACE. Parameters not listed take the default values of scikit-learn and XGBoost.

\begin{table}[b!]
\centering
\caption{Hyperparameters of the outcome predictor, the reference models, and FACE.}
\label{tab:hyperparameters}
{\small
\begin{tabularx}{\columnwidth}{lX}
\toprule
Model                                  & Hyperparameters                                                                                  \\
\midrule
Linear Regression                      & ---                                                                                              \\
Ridge Regressor ($\hat{H}_p$)          & $\alpha = 1.0$                                                                                   \\
Random Forest                          & 500 trees, maximum depth 10                                                                      \\
XGBoost                                & 300 trees, maximum depth 6, learning rate 0.01, row subsampling 0.9, column subsampling 0.9      \\
MLP                                    & hidden layers (256, 64), ReLU activation, Adam optimizer, 2000 maximum iterations                \\
FACE                                   & $k = 10$ nearest neighbors, Gaussian kernel density with Scott's bandwidth rule                  \\
\bottomrule
\end{tabularx}}
\end{table}

\section*{References}
\vspace{-1.5em}
\bibliographystyle{IEEEtran}
\bibliography{main}

\end{document}